# THE CHARACTER THINKS AHEAD: CREATIVE WRITING WITH DEEP LEARNING NETS AND ITS STYLISTIC ASSESSMENT

Roger T. Dean and Hazel Smith, austraLYSIS and Western Sydney University, NSW 2751, Australia.  Email: <roger.dean@westernsydney.edu.au>.

**Abstract**

We discuss how to control outputs from deep learning models of text corpora so as to create contemporary poetic works. We assess whether these controls are successful in the immediate sense of creating stylometric distinctiveness. The specific context is our piece *The Character Thinks Ahead* (2016/17); the potential applications are broad.

Our recent piece *The Character Thinks Ahead,* is focused on the computerized generation of creative writing using deep learning neural nets: the competitive and progressive learning process behind this becomes a key topic of the text. The piece contains many other elements that interact with the text generation to create a multi-layered whole: these include pre-composed poetic text, both screened and performed (by author HS), and improvised and composed sound. Here we discuss the piece, the processes that comprise deep learning, and in particular our attempts to create and test for distinctive stylistic outputs within the computer-generated text.

*The Character Thinks Ahead* knits together visual, sonic, linguistic and literary elements that all interact with each other (see Figure 1, and the supplemental video of the piece). Of the three dynamically rolling columns of text in the upper part of the screen, the middle presents three pre-composed poetic texts that suggest ideas, feelings and contexts to do with war, hierarchy and competition respectively. The two columns on either side display text generation using deep learning nets (described later): in the left column the text is generated by character [1], in the other it is generated by word. In the bottom part of the screen there are also three distinct elements to the display. An animated word cloud in the middle highlights features of the ongoing texts. To the left of it is a dynamic spectral visualization of a (pre-recorded) rendering of the live speech: this is live-transformed to provide a sonic output visualized spectrally on the right. Besides the visual elements, the live speech and the live sonic transformation, there is also pre-formed sound — composed and improvised by author RTD with further improvised contributions from members of our ensemble austraLYSIS, Sandy Evans (saxophone) and Greg White (computer).

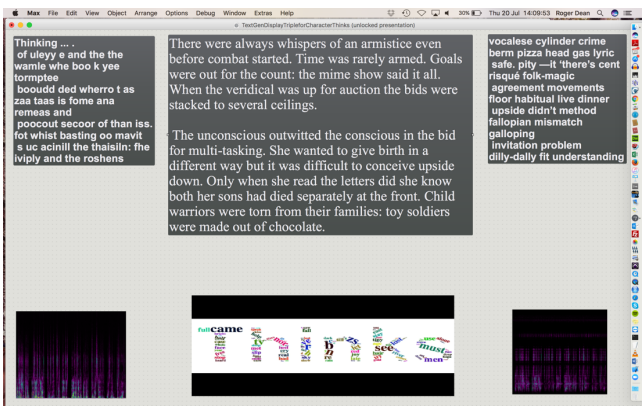

Fig. 1. A screenshot from a performance of *The Character Thinks Ahead*. (© The authors.)

The spoken text plays on different senses of the word character: character as part of a word, as a register of behavior or as a fictional being. It also relates to "thinking ahead", central to the predictive aspects of deep learning (or even "thinking a head"). Ideas of competition between word and character, which are a feature of the spoken text, are also a feature of the process of text generation. These ideas are further explored in the screened pre-composed poetic texts (middle text panel).

## Computational Creativity: Intent and Assessment

Our main computational purpose was to develop deep learning models of two corpora of poetic texts (one we term the Gutenberg, mainly last century poetry, the other the Smith corpus, hereafter termed the 'HS corpus' [2]). The models (below) aim to predict the next word following an input sequence. But rather than allow this prediction to reflect solely the corpus features, we wanted to transform the predictions by the choice of input sequence (normally from new HS texts) and by the mode of sampling the predictions: a deep learning model outputs an estimate of the probability for each word in its vocabulary that it will be the next word. Normally the most probable word is chosen but we explored alternatives. We used computational stylometric assessments to determine whether we succeeded in transforming the outputs so that they were distinct from both the corpus and from the new HS texts.

A subsidiary intent in the piece was to use the deep learning models at various stages during training so as to reflect the gradual improvement in their predictions. For example, when a model is used that predicts the next character (space, punctuation or letter) of a sequence, at the outset simply the most frequent individual elements form the predictions (many repeated letters and few words). Gradually during training letters assemble and increasingly constitute recognizable words. This process is reflected in the left-hand text display box of the piece. To a lesser degree, the analogous learning process in word prediction is reflected in the right-hand top text display box.

Our deep learning nets were Recurrent Neural Nets (using Long Short Term Memory or Gated Response Unit nodes) and Convolutional Neural Nets (CNN) [3]. The names of the nets reflect the specific computational nature of their nodes and overall structure. In essence, a neural net is a set of interconnected nodes, each of which computes a transformation (mostly non-linear) of its cumulated inputs. The nodes are layered with various degrees of interconnection: a first layer receives the input and the final layer gives a prediction of the output. Deep learning nets are only 'deep' in the sense of having large numbers of nodes and also often layers. Here each input in the first layer was a sequence of 30 words taken from the corpus, or a 'seed' of 30 words for the later use of a trained model for generation, and the final layer was the output, the single next word predicted to follow the input sequence. Our CNNs were dilated, meaning that as required for sequence predictions, input n might influence not only output n, but also outputs 2,4, 8, 16, 32 later (as chosen). The nets learned (progressively adjusting the parameters of each node during exposure to a 'training' set of word sequences) to optimally predict the next word of a previously unseen 'test' set.

An example of input and output from a trained net (that is no longer learning) follows. Note that the initial 30-word seed predicts the next word, but then the first seed word is dropped and the original words 2-30 — plus the newly predicted word as terminal — predict the second generated word, and so on. No natural language processing or text selection is applied.

**Input** (a 30 word sequence from an HS text, taken from our earlier real-time text generation piece *Instabilities II* [4]): *the place of no return. your rights have been withdrawn. don't touch the screen! judged not by the colour of her skin but the content of her character. drudgery. the.*

**Output** (Part of a 1000 word sequence): *he's cross-weathering re-de-po-e-pa long, smaller, rejigging richer illusions somersault desiccation     short e. words, kids debut asynchrony. grey. jumble, heroism voicemail came     polished. names urges: aside hot hanging starved. aggression chap     parliaments ocean, four sparingly, feeling     route. palestinians learn thieving. barely payoff, conjoin surfing her, w hover     sinking calama assume tremble. successful dr. invite laws     thunder laced-up, lies calama payoff interfering edges laughter list-making laws perverse, better.     kept sms cut specious at quoi     arms. pleas yoga-time him. roberts others. dark out-of-orbit aggression diagnosis legs obliquely sort alone gaze stockpiles theory something, leant report     result     her. trees succession remix mourn cooled. milan.     room-temperature détente.*

This output has only been post-processed by the addition of spaces between the words, to encourage different modes of reading. We make no claim as yet for the literary quality of the model outputs, though we find them resonant when juxtaposed with composed texts. We intend to evaluate such outputs empirically later, and to use a comparative context of experimental 20th/ 21st century poetry. This complex task should be done by both experts and non-experts and will require removing references to computational generation: see [5] for comparative discussion of evaluation issues in short-term music generation. Meanwhile, we have assessed whether we have achieved distinctive outputs, using computational stylometry (based primarily on relative word frequencies using the R package 'stylo' for the analysis).

Figure 2 summarizes one such analysis in the form of a cluster diagram. This shows relationships amongst a set of generated texts and the three composed texts of HS that are progressively displayed in the upper central text panel of the piece. In this case, for a particularly stringent test, the model was trained on the Smith corpus text and seeded with extrinsic HS text. The cluster analysis groups texts and text sets, according to the distances between their relative word frequency patterns (shown as the value on the horizontal axis at which they are joined by a vertical line). The first salient variable assessed in this particular analysis was the influence of seeding the text generation with randomly selected sequences of previously unseen HS text versus self-seeding it from the HS corpus. The second was the influence of the diversity (D values in the figure) used in sampling model predictions, where higher values of D, usually called 'temperature', allow greater diversity of output from each probability distribution for the next word. Figure 2 confirms that seeding creates distance between self-seeded and (new) HS-seeded outputs, and that as D increases, so distances increase systematically. The grouping of the three *Character Thinks Ahead* texts at the bottom is vastly distinct (distance 2.5) from the grouping comprising all the generated texts, while the three texts themselves are the most homogeneous group (lowest distance measures). This again suggests that the generative machinery is not simply reproducing the style of the learned corpus. When similar stylometry is performed on the outputs of the 'Gutenberg' (+HS) corpus model, very similar conclusions are reached, and the composed corpora group is stylistically separate from the generated texts. Interestingly, when word 'preference' and 'avoidance' in the 'Gutenberg' CNN generated texts is investigated, some of the words most preferred as a consequence of seeding with HS texts do not derive from those seeding texts: another confirmation that we created stylometric differences between the outputs of the model when self-seeded, when seeded with HS texts, and in response to the diversity parameter.

We are investigating further mechanisms by which sampling diversity (and coherence) can be controlled for the purposes of creating stylistic difference. In addition, issues such as the rate of change of computationally measured sentiment (enhanced in the seeded outputs in comparison with the Gutenberg corpus self-seeding), topic fluxes or other literary qualities will be of interest. Some of these assessments lead to cognitive predictions: for example, that a continuous measure of affect perceived by a reader should vary more when there are increased fluxes in measured sentiment, because the reader may be responsive to those fluxes. Such issues can be assessed with carefully designed perceptual methodologies, complementing reader assessments of liking, stylistic diversity and in depth literary analysis. These will need to take account of the ongoing literary context, just as real-time computational music generation has to take account of the musical context: computational generation may be most useful in conjunction with simultaneous human performance, but needs to be assessed with and without this [5].

**Fig. 2. Cluster Analysis of Outputs from a model trained on the HS corpus and of three separate texts by Hazel Smith. D indicates sampling diversity; 'SmithSeed' indicates seeding by new HS text, otherwise the text self-seeds.**

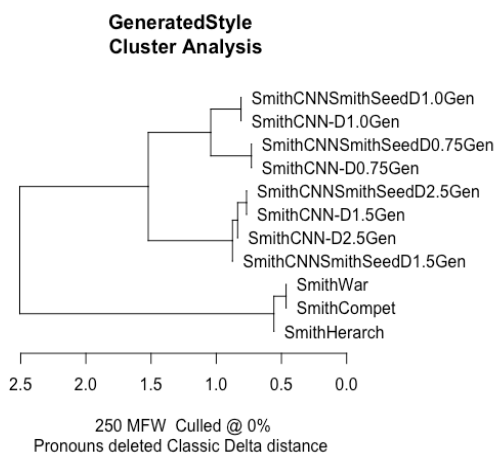